\documentclass[runningheads]{llncs}
\usepackage{graphicx}

\usepackage{tikz}
\usepackage{comment}
\usepackage{amsmath,amssymb} 
\usepackage{color}
\usepackage{multirow}
\usepackage{caption} 
\usepackage{cuted}
\usepackage{enumitem}
\usepackage{xcolor}
\usepackage{pifont}
\newcommand{\cmark}{\ding{51}}%
\newcommand{\xmark}{\ding{55}}%
\usepackage{wrapfig}

\usepackage[pagebackref,breaklinks,colorlinks]{hyperref}

\begin{document}
\pagestyle{headings}
\mainmatter
\def\ECCVSubNumber{4685}  

\title{GALA: Toward Geometry-and-Lighting-Aware Object Search for Compositing} 

\titlerunning{GALA}
%
\author{Sijie Zhu\inst{1,\dagger}, Zhe Lin\inst{2}, Scott Cohen\inst{2}, Jason Kuen\inst{2}, Zhifei Zhang\inst{2}, Chen Chen\inst{1}}
\authorrunning{Zhu et al.}
%
\institute{Center for Research in Computer Vision, University of Central Florida \and
Adobe Research\\
\email{sizhu@knights.ucf.edu,\{zlin,scohen,kuen,zzhang\}@adobe.com,chen.chen@crcv.ucf.edu}}
\maketitle
\let\thefootnote\relax\footnotetext{$\dagger$ This work was done during the first author’s internship at Adobe Research.}
\begin{abstract}
Compositing-aware object search aims to find the most compatible objects for compositing given a background image and a query bounding box. Previous works focus on learning compatibility between the foreground object and background, but fail to learn other important factors from large-scale data, i.e. geometry and lighting. To move a step further, this paper proposes GALA (Geometry-and-Lighting-Aware), a generic foreground object search method with discriminative modeling on geometry and lighting compatibility for open-world image compositing. Remarkably, it achieves state-of-the-art results on the CAIS dataset and generalizes well on large-scale open-world datasets, i.e. Pixabay and Open Images. In addition, our method can effectively handle non-box scenarios, where users only provide background images without any input bounding box. A web demo (see supplementary materials) is built to showcase applications of the proposed method for compositing-aware search and automatic location/scale prediction for the foreground object. 
\end{abstract}

\section{Introduction}
\label{sec:intro}
Compositing-aware object search/retrieval \cite{cais} aims to find suitable source images for compositing \cite{niu2021making,zhang2021deep}. Specifically, given a background image and a bounding box indicating the compositing location, the objective is to retrieve compatible foreground objects from a large reference database, so that the composite image appears realistic.  Harmonization \cite{jiang2021ssh,zhu2015learning,tsai2017deep,azadi2020compositional,lin2018st} is then applied to adjust the color and edge pixels, but it is extremely challenging to automatically adjust the semantics, lighting, or geometry of foreground objects. Although recent relighting method [19] can generate realistic lighting change for human portraits, it does not tackle general object categories and the required 3D reconstruction is not always available. Therefore, the quality of the final composite image highly depends on the performance of foreground retrieval system, and a good system should be aware of semantics, lighting, and geometry of foreground objects. 

\begin{figure}[!htbp]
    \centering
    \vspace{-0.7cm}
    \includegraphics[width=\linewidth]{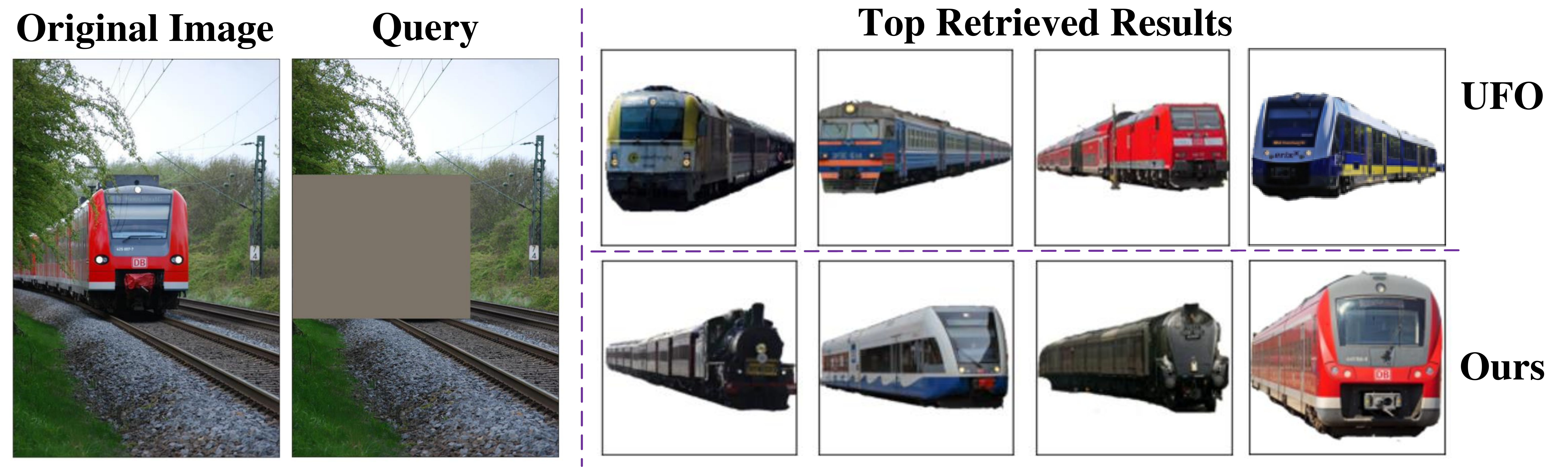}
    \vspace{-0.3cm}
    \caption{Comparison between state-of-the-art method (UFO \cite{ufo} in the first row) and the proposed geometry-and-lighting-aware search (second row). The retrieved objects in the first row do not respect the geometry of the background scene, while results of the proposed method have better geometry compatibility. }
    \label{fig:front2}
\end{figure}

\begin{figure}[!htbp]
    \centering
    \vspace{-1.4cm}
    \includegraphics[width=.95\linewidth]{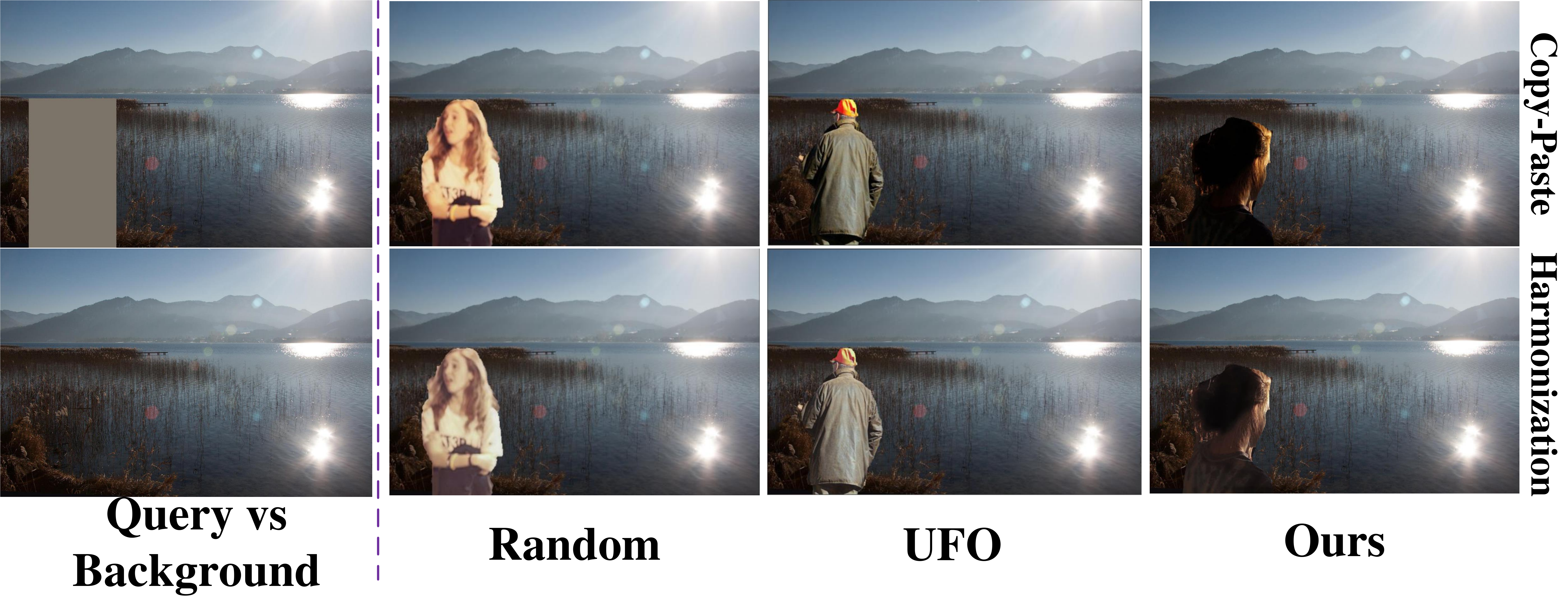}
    \vspace{-0.3cm}
    \caption{Comparison between state-of-the-art method (UFO \cite{ufo}) and the proposed method. Objects retrieved by ``Random" and ``UFO" do not have good lighting compatibility even after harmonization. Our method better respects the lighting condition (light coming from front-right direction), and thus the final composite image is more realistic.}
    \label{fig:front3}
\end{figure}

Early work ~\cite{cais} on foreground retrieval follows a constrained setting, which requires the user to specify the object category for retrieval. Specifying the category sets a limit on the search space, thus preventing the system from recommending diverse sets of objects. Later, UFO~\cite{ufo} proposes an unconstrained search method, i.e. objects from all categories are considered as candidates for retrieval. The unconstrained setting is closer to real-world scenarios, which require a large and diverse foreground object gallery to satisfy different users. However, UFO \cite{ufo} focuses on finding semantically compatible foreground object and does not explicitly model lighting and geometry, which are critical factors for making object compositing realistic, as shown in Figs.~\ref{fig:front2} and \ref{fig:front3}. More recent work \cite{li2020interpretable,wu2021fine} either explores additional annotation \cite{li2020interpretable} for constrained setting, or only focuses on indoor furniture \cite{wu2021fine} with fine-grained sub-categories. Therefore, how to encourage awareness on geometry and lighting is still unclear for general unconstrained foreground retrieval. One way to consider these factors is to explicitly estimate the 3D geometry and lighting of the scene, 
but it is extremely challenging to obtain realistic training data with 3D labels to do so. Instead, we aim to build a discriminative model which is sensitive to lighting and geometry mismatches using real-world image datasets, leading to a more generalizable and scalable solution.

In this paper, we propose a novel Geometry-And-Lighting-Aware (GALA) foreground object search system, which aims to retrieve objects that are compatible in terms of semantics, geometry, and lighting. Specifically, we design a model consisting of a foreground object encoder and background encoder, such that only the matching pairs of foreground object and background are closer to each other in the embedding space. To encode geometry and lighting sensitivity, we generate negative samples by augmenting the same objects with very different geometry and lighting conditions through homography transformation with left-right flip and non-linear illumination modification on foreground objects during training. Contrastive learning is then applied to push the transformed foreground object far away from the original one in the embedding space. However, the semantic compatibility of retrieved objects could degrade significantly if we train the foreground network jointly with the background network. Thus, we introduce an alternating training strategy to maintain semantic compatibility, while learning to respect geometry and lighting. GALA can also be extended to handle non-box scenarios, where the users only provide the background image without any bounding box or text input. Our model automatically retrieves foreground objects and predicts the best location and scale for compositing. 

Furthermore, we perform experiments and validate GALA for open-world datasets with an arbitrarily large number of categories, targeting real-world unconstrained applications. Previous works are mostly implemented on medium scale datasets ($<100,000$ objects) with limited categories, e.g. MSCOCO \cite{lin2014microsoft} only has 80 object categories. In contrast, we conduct experiments on large-scale real-world datasets, i.e. Pixabay \cite{pixabay} and Open Images \cite{openimages}, which contain significantly more categories and images with diverse contents. 
We show comparisons of various settings between the proposed method and previous works in Table \ref{tab:setting}. Our contributions are summarized as follows:
\setlist{nolistsep}
\begin{itemize}[noitemsep,leftmargin=*]
    \item A novel compositing-aware search method that discriminatively models geometry and lighting with contrastive training.
    \item An alternating training strategy to address the challenge of losing semantic compatibility when learning the foreground network in an unconstrained setting. 
    \item Extensive experiments to demonstrate significant improvements over previous works on the CAIS dataset, as well as a new large-scale open-world dataset, i.e. Pixabay. 
    \item An extension of the method for non-box conditioned scenarios which are never considered in previous works. 
\end{itemize}

\begin{table}[!htbp]
    \centering
    \caption{Comparison between our method and previous works on settings.}
    \begin{tabular}{l|c|c|c|c|c}
    \hline
    
    \hline
         & Zhao \cite{cais} & UFO \cite{ufo} & Li \cite{li2020interpretable} & Wu \cite{wu2021fine} & Ours \\
         \hline
    One model for all & \cmark & \cmark & \cmark & \xmark & \cmark \\
      Unspecified Class & \xmark & \cmark & \xmark& \cmark & \cmark \\
      Non-box Scenarios & \xmark & \xmark & \xmark & \xmark & \cmark \\
      Large-scale Dataset & \xmark & \xmark & \xmark & \xmark & \cmark \\
      Open-world Setting & \xmark & \xmark & \xmark & \xmark & \cmark \\
         \hline
         
         \hline
    \end{tabular}
    
    \label{tab:setting}
    \vspace{-0.3cm}
\end{table}



\section{Related Work}
\label{sec:related}
\noindent\textbf{Search for Compositing.} The idea of searching for foregrounds to insert into a new background is first proposed by Lalonde et al. \cite{lalonde2007photo}, by explicitly estimating the 3D geometry of foreground objects, and the lighting map of background scenes. Zhao et al. \cite{cais} propose a learning-based compositing-aware search method, where the users provide a specific class as text input. The positive samples are augmented based on shape and semantics. UFO \cite{ufo} assumes that the user does not specify any category, which is the most relevant setting to our work, but it only focuses on the semantic compatibility and ignores lighting and geometry. It trains a discriminator to distinguish real/fake object for a certain background and selects candidate positive samples for each background image. Li et al. \cite{li2020interpretable} manually annotates multiple attributes as additional information to determine “Interpretable Foreground Object” with given categories. Wu et al. \cite{wu2021fine} propose a teacher-student framework for fine-grained indoor categories, which adopts multiple pre-trained features for foreground objects as teacher and trains the background network with KL-divergence. None of them learn discrimitive features on geometry and lighting for general unconstrained foreground search. \\
\textbf{Object placement.} A thread of works \cite{lee2018context,zhang2020learning,li2019putting} focus on object placement along with adversarial training, which predicts locations and scales to insert an object. Lee et al. \cite{lee2018context} train a spatial transformer network to predict the location for placement and a shape mask to guide generation. Zhang et al. \cite{zhang2020learning} train a location-scale prediction model using inpainted pure background images. Li et al. \cite{li2019putting} first find candidate locations in indoor scenes, then predict the best human pose for a specific location. These methods either focus on street scenes with limited object categories, e.g. person, car, or indoor scenes with only human pose joints. Our location-scale prediction deals with a more challenging setting with general objects from diverse scenes by directly adopting our retrieval model. \\

\vspace{-0.5cm}
\section{Method}
In this section, we introduce GALA, Geometry-and-Lighting-Aware foreground object search for compositing. We first formulate the problem and describe training data generation in Sec. \ref{sec:problem}. Then we present the details of contrastive learning with self-transformations and the alternating training strategy in Sec. \ref{sec:contrastive} and \ref{sec:alternative}. Finally, we show how GALA handles non-box scenarios in Sec. \ref{sec:location}.

\subsection{Problem Statement}
\label{sec:problem}
Given a background image $I_{b}$ with a bounding box $(l,r,w,h)$, our objective is to retrieve a set of most compatible foreground object images $\{I_{f}\}$, so that realistic composite images can be generated with simple harmonization techniques. Our framework aims to learn an embedding space so that compatible images are close to each other in this space and the ranking can be obtained by simply computing the cosine similarity or euclidean distance. Since foreground and background images have very different distributions and appearances, we use two different encoder networks $N_{b}, N_{f}$ to generate the embedding features for $I_{b}, I_{f}$, respectively. During training, the only available annotation is semantic segmentation mask for each object. We use the mask to crop out the object as $I_{f}$. Then $I_{b}$ is generated by applying a rectangle mask on the image covering the whole object. Since there is no manual annotation on positive and negative samples, we consider $I_{b}$ and $I_{f}$ generated from the same image as a positive sample to each other, while other image pairs are considered as negative samples. With a hard-margin triplet loss \cite{triplet}, the optimization can be formulated as $\operatorname*{arg\,min}_{N_b,N_f} \mathcal{L}_{t}$, where
\begin{equation}
\mathcal{L}_{t}=[S(N_{b}(I_{b}),N_{f}(I_{f}^{-})) - S(N_{b}(I_{b}),N_{f}(I_{f}^{+}))+ m]_{+}.
\label{eq:1}
\end{equation}
\noindent Here $S$ means cosine similarity, and $[.]_{+}$ denotes the hinge function. $m$ is the margin for triplet loss and $I^{+}_{f}, I^{-}_{f}$ indicate the positive and negative foreground objects for a given $I_{b}$. 

\subsection{Contrastive Learning with Self-transformation}
\label{sec:contrastive}
The standard loss in Eq. \ref{eq:1} only considers the contrastive information between the original foreground object and other objects, which mainly focuses on semantic information. We argue that a Geometry-and-Lighting-aware system should be able to tell the difference if the foreground object has exactly the same semantics with different lighting and geometry conditions. We thus perform transformation on the original positive foreground object so that it can be considered as negative. 
\begin{figure}[!htbp]
\vspace{-0.5cm}
    \centering
    \includegraphics[width=0.7\textwidth]{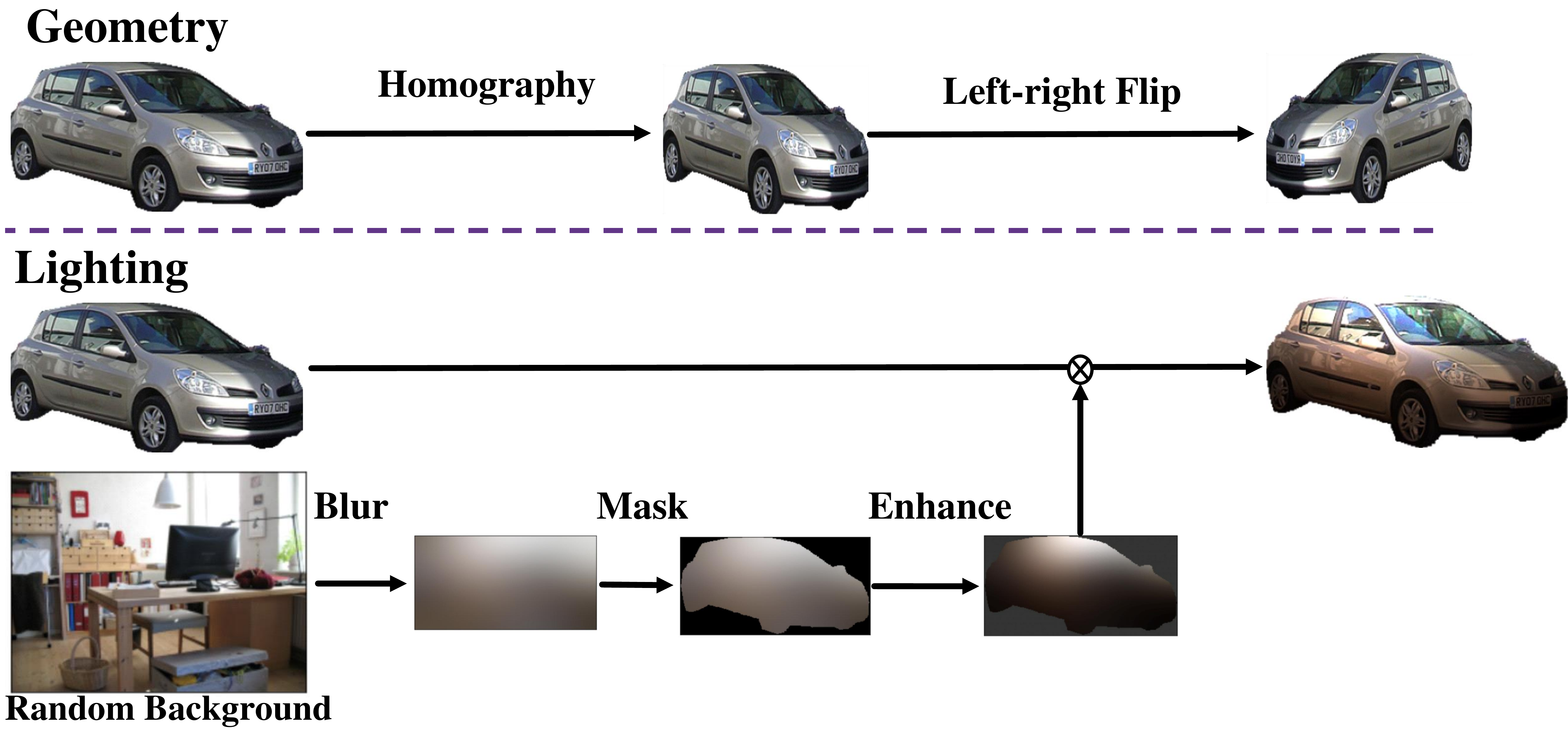}
    \vspace{-0.3cm}
    \captionof{figure}{Pipeline of generating lighting and geometry transformations used in contrastive learning.}
    \label{fig:transformation}
    \vspace{-0.5cm}
\end{figure}
As shown in Fig. \ref{fig:transformation}, the transformed foreground objects have very different lighting or geometry conditions, which do not match with the background image anymore. We consider the transformed object image $I_{f}^{t}$ as negative and formulate another triplet loss (Fig. \ref{fig:contrastive}):
\begin{equation}
\mathcal{L}_{c}=[S(N_{b}(I_{b}),N_{f}(I_{f}^{t})) - S(N_{b}(I_{b}),N_{f}(I_{f}))+ m]_{+}.
\label{eq:2}
\end{equation}

The final loss is given by $\mathcal{L}=\mathcal{L}_{t}+\mathcal{L}_{c}$ from Eqs. \ref{eq:1} and \ref{eq:2}. In Fig. \ref{fig:transformation}, we show the pipeline of self transformations. For geometry transformation, we first apply random homography transformations, then left-right flip is applied with a $50\%$ probability. For lighting transformation, we first find a random background image and apply Gaussian blur with a large radius, e.g. 100. The blurred map is then resized with interpolation to the size of foreground object, and masked with the segmentation mask. Finally, we enhance the variance of the lighting map with an exponential function so that the largest value of the map is $5$. In this way, we get a non-linear illumination map which highlights a random region in the object image, and it is multiplied by the original object image to generate the final transformed image. Although the lighting may not be as realistic as 3D relighting \cite{pandey2021total}, it is enough to be considered as a negative sample with very different illuminations.

\hspace{-0.6cm}
\begin{minipage}[t]{0.55\linewidth}
    \centering
    \vspace{-0.1cm}
    \includegraphics[width=0.75\textwidth]{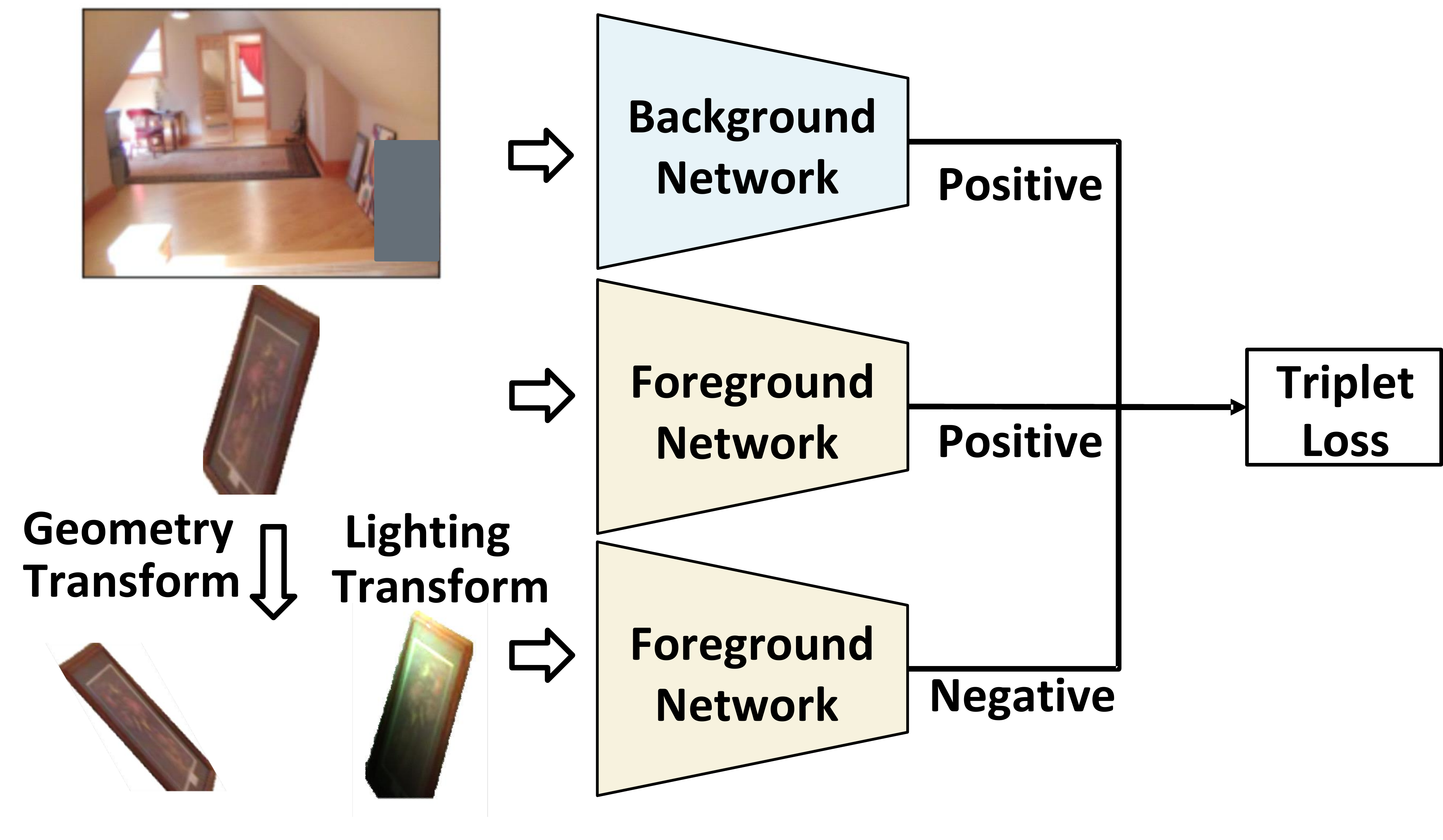}
    \vspace{-0.4cm}
    \captionof{figure}{Contrastive learning with self-transformations.}
    \label{fig:contrastive}
\end{minipage}
\hspace{0.2cm}
\begin{minipage}[t]{0.4\linewidth}
\captionof{table}{Performance of different training strategies on CAIS.}
    \centering
    \begin{tabular}{l l}
    \hline
    
    \hline
    & mAP \\
    \hline
    Fix Foreground & 29.10  \\
    Direct Training & 17.93  \\ 
    Aug & 23.45  \\ 
    Aug + Alternating & \textbf{31.20} \\ 
    \hline
    
    \hline
    \end{tabular}
    \label{tab:training}
\end{minipage}


\subsection{Alternating Training}
\label{sec:alternative}

We notice that previous works \cite{ufo,wu2021fine} generally use pre-trained weights for foreground network $N_{f}$ and freeze its parameters during training (denoted as ``Fix Foreground"), which means $N_{f}$ cannot learn from data. Since data augmentation like left-right flip is adopted for pre-training on ImageNet \cite{deng2009imagenet}, the feature would be invariant to the left-right flip, which could change the geometry and direction of lighting. Therefore, the pre-trained weights are not discriminative to lighting and geometry changes, which will be further demonstrated in Sec. \ref{sec:ablation}. 
  
\hspace{-0.6cm}
\begin{minipage}[b]{0.4\linewidth}
\includegraphics[width=\textwidth]{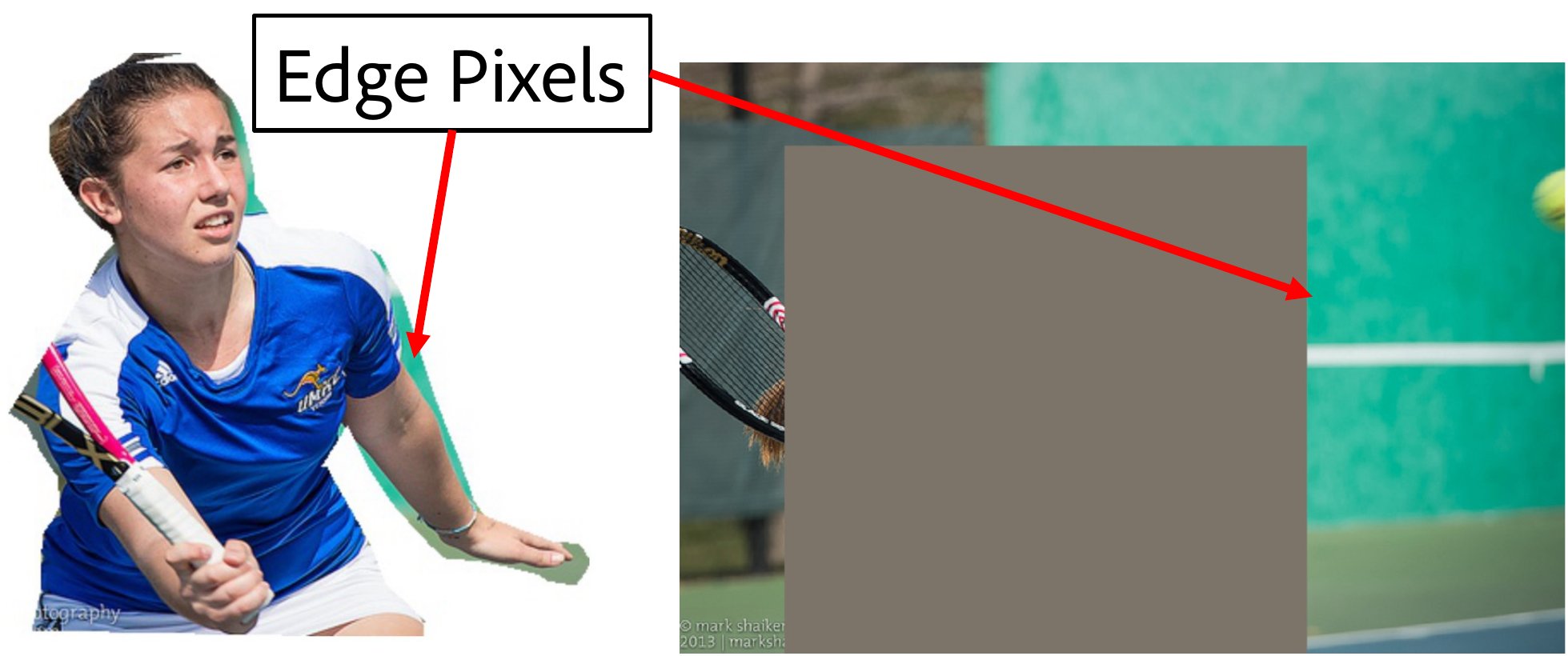}
\captionof{figure}{Imperfect segmentation mask and edge pixel example.} 
\label{fig:edge}
\end{minipage}
\hspace{0.2cm}
\begin{minipage}[b]{0.57\linewidth}
\includegraphics[width=\textwidth]{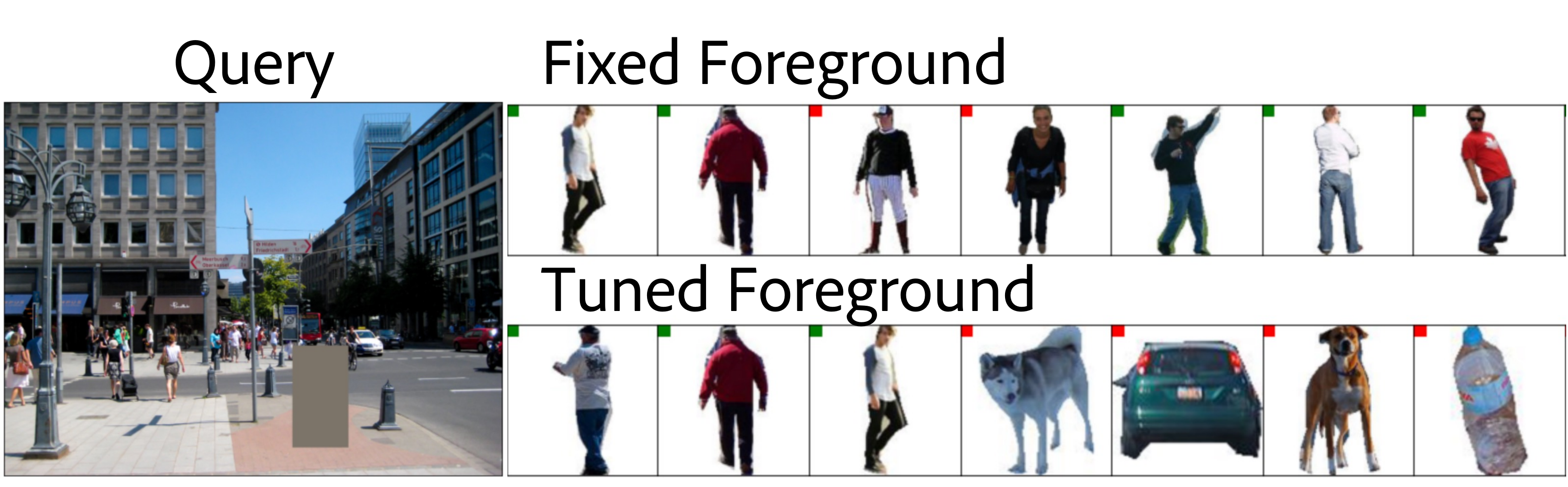}
    \captionof{figure}{An example of poor compatibility on semantics when training foreground network.}
    \label{fig:semantics}
\end{minipage}

\indent To learn geometry-and-lighting-aware representation, $N_{f}$ must be learnable. However, performance drops significantly if $N_{f}$ and $N_{b}$ are directly trained with Eq. \ref{eq:1} (denoted as ``Direct Training" in Table \ref{tab:training}), partly due to the imperfect segmentation mask, as some annotations (MSCOCO) are based on polygon. When cropping with an imperfect mask (Fig. \ref{fig:edge}), some edge pixels of the object are actually background pixels, which are very likely to be the same as other background edge pixels. Direct training may optimize the model to match the edge pixels without compatibility on semantics, as the edge pixel is a very strong cue for positive pairs. This issue could be tackled by additional mask augmentations (denoted as ``Aug") to prevent the model from using such cue. We randomly erode the foreground mask and extend the background mask so that the cue becomes random. Examples are included in the \textcolor{blue}{supplementary material}. ``Aug" significantly improves the performance over direct training.

\indent However, the performance is still much lower than using fixed foreground network. Another issue is that the foreground object images have very different appearances and distributions from regular images. And the ``Direct Training" model $N_{f}$ does not respect semantics very well as compared with using ImageNet pre-trained weights. As shown in Fig. \ref{fig:semantics}, ``tuned foreground" retrieves irrelevant
\begin{wrapfigure}{r}{0.5\linewidth}
    \centering
    \vspace{-0.8cm}
    \includegraphics[width=0.5\textwidth]{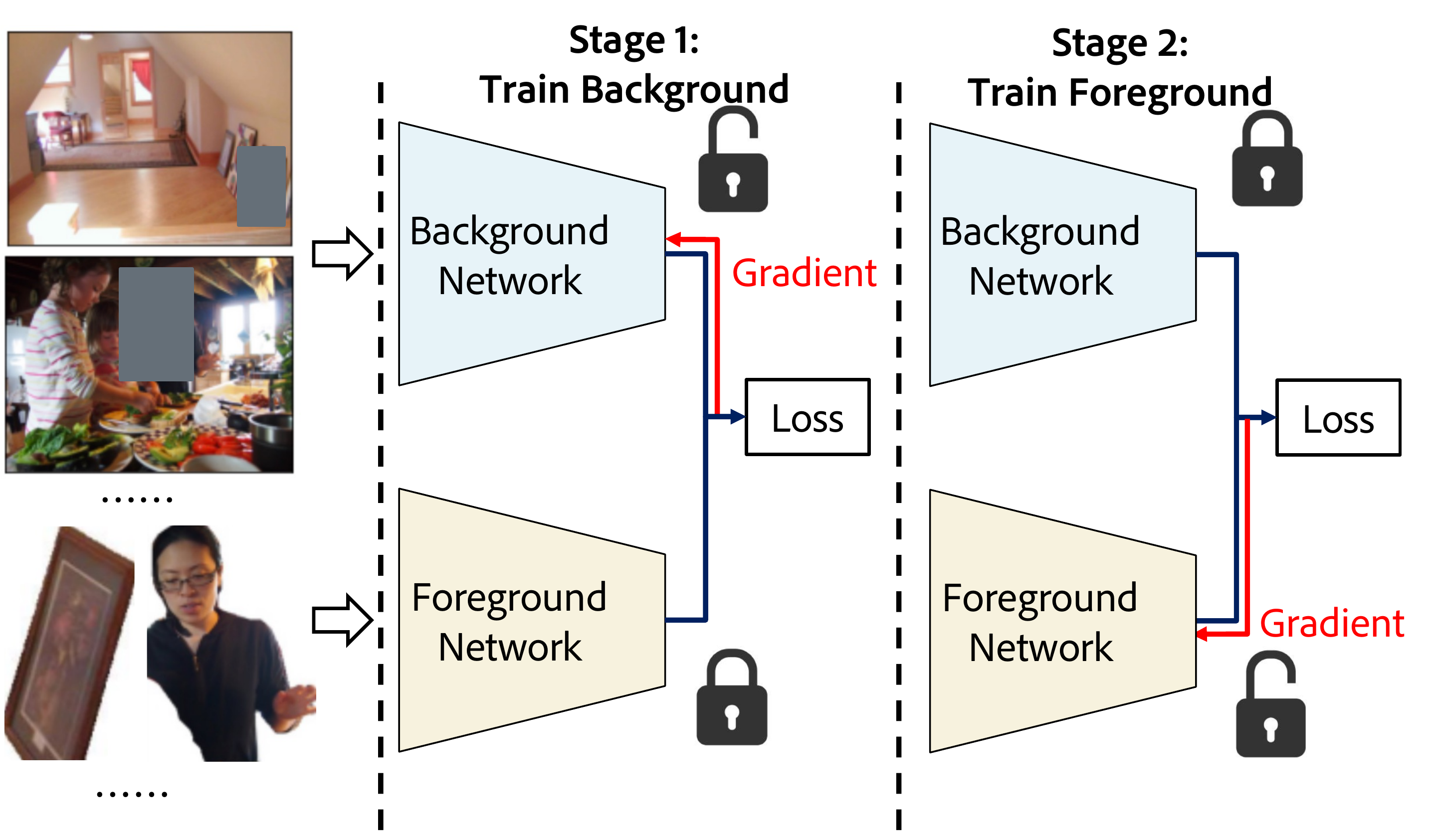}
    \vspace{-0.7cm}
    \caption{The proposed alternating training strategy.}
    \vspace{-1.0cm}
    \label{fig:alternative}
\end{wrapfigure}
categories for a sidewalk query, e.g. bottle, car. Our solution is to alternatively train $N_{b}$ and $N_{f}$ as shown in Fig. \ref{fig:alternative}. Since there is only one network trained in one stage, the embedding features of the trained network will not have much drift. In this way, our method maintains semantic information in foreground feature, while allowing $N_{f}$ to learn from data for other factors, i.e. lighting and geometry. 




\subsection{Extension to Non-box Scenarios}
\label{sec:location}
\begin{wrapfigure}{r}{0.55\linewidth}
    \centering
    \vspace{-0.8cm}
    \includegraphics[width=0.55\textwidth]{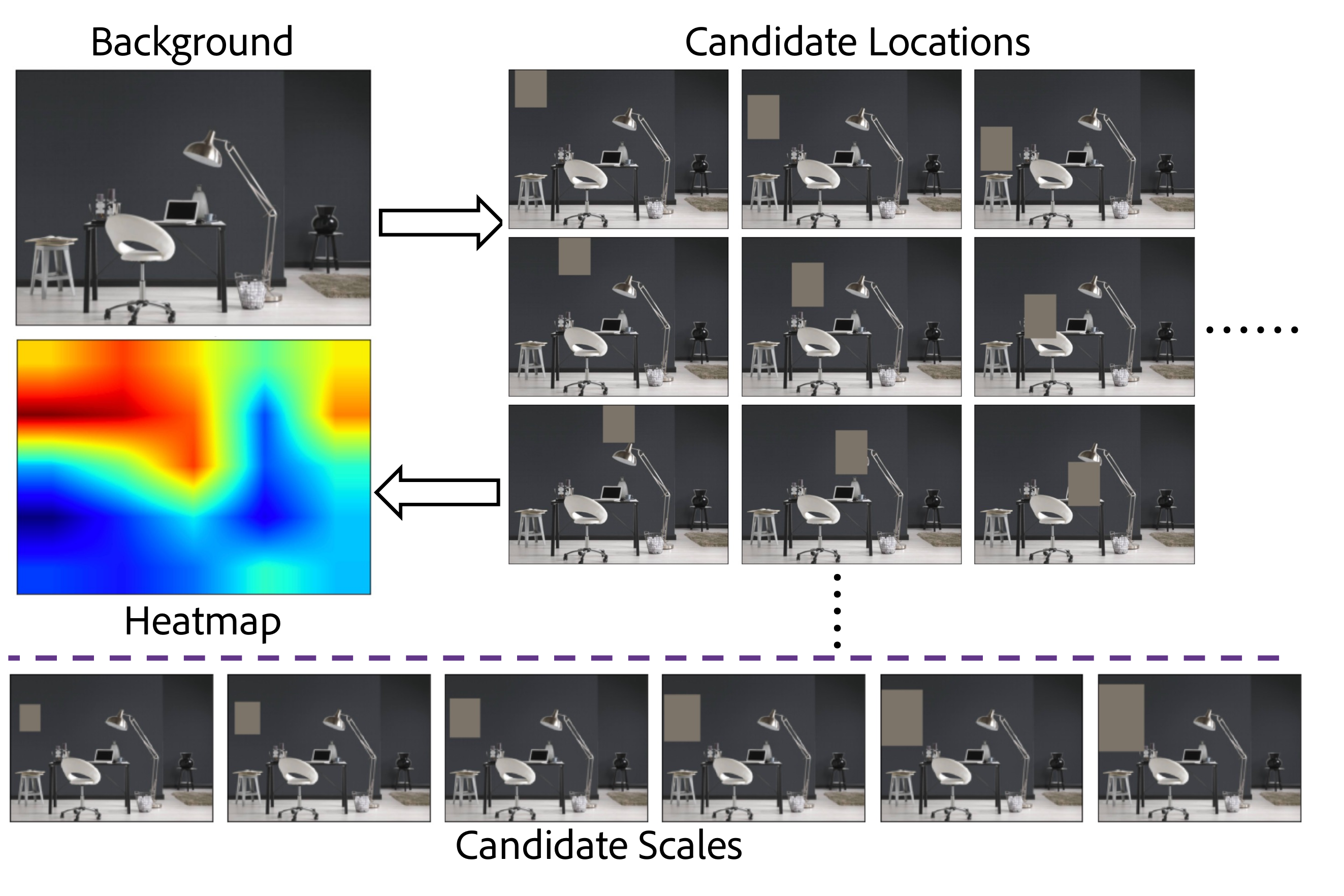}
    \vspace{-0.7cm}
    \caption{Example of location/scale prediction.}
    \vspace{-0.5cm}
    \label{fig:nonclick}
\end{wrapfigure}
When the user only provides a background image without any bounding box, we adopt the following random seed and greedy search algorithm to retrieve matching foreground objects. We first sample multiple random locations. Each location is assigned with multiple bounding boxes with different aspect ratios and scales. We retrieve foreground objects based on each of the bounding boxes and re-rank all the results based on cosine similarity. We select the best one as the default object, then assign an initial bounding box with the same aspect ratio as the object. The area of the initial box is empirically assigned as $1/25$ of the query image. Then we apply $k \times k$ grid of locations (Fig. \ref{fig:nonclick}) with initial bounding box to cover the query image in a sliding window manner. By computing the similarity score between the object and background with all the possible boxes, the box with the highest score is considered as the best location. A location heatmap is generated by interpolating the $k\times k$ score matrix to the size of query image. One can always increase $k$ or apply further refinement with smaller stride to improve the location. The scale is then selected by applying a range of scale ratios on the initial box at the best location, as shown in Fig. \ref{fig:nonclick}.

\vspace{-0.1cm}
\section{Experiment}
We first conduct experiment on CAIS \cite{cais}, which is specifically annotated for compositing-aware search. Then we generate two large-scale datasets based on Pixabay \cite{pixabay} and Open Images \cite{openimages} to demonstrate our model's generalization ability on open-world setting.\\ 
\noindent\textbf{CAIS:} The dataset is generated by selecting 8 popular categories from three datasets, i.e. MS-COCO \cite{lin2014microsoft}, PASCAL VOC 2012 \cite{everingham2010pascal}, ADE20K \cite{zhou2017scene}. The training set has $86,800$ background images, and the original foreground object in each image is considered as ground-truth. The evaluation set has $80$ manually selected background images and query bounding box to insert objects. About $16\sim 140$ compatible foreground objects are annotated for each background image.\\
\textbf{Pixabay:} ``pixabay.com" is a stock image website containing tons of high-quality photos which are perfect for composing. The images are highly diverse, free to use, and substantial in number. We first collect about $928,018$ images, then apply object detection \cite{tian2019fcos} and segmentation \cite{lee2020centermask} to generate foreground objects and background images based on masks. In total, we get $5,771,912$ foreground objects and $928,018$ background images. But some objects are not well segmented or have extremely small size which is not likely to be used for composting. Background images with overly large box is also not suitable as there is nothing left to tell what should be here. Therefore we only keep images with high confidence score (e.g. $>0.6$) and proper bounding box size (e.g. box area in $5\sim 50\%$ of the area of whole image). Finally, we get $833,964$ foreground and background pairs with $914$ non-zero categories. They are then randomly split into training/evaluation set with portion of $90\%/10\%$.\\
\textbf{Open Images:} Open Images originally contains about $9$ million images with $9,605$ trainable classes, thus is perfect for open-world evaluation. Up to $2.8$ million objects from $350$ categories are annotated with segmentation masks. We originally select $944,024$ images with $2,686,666$ objects, then apply the same filtering procedure as Pixabay based on box size only, as the mask is annotated. In total, we keep $1,374,344$ background and foreground object pairs in our experiments. They are then randomly split into training/evaluation set with portion of $90\%/10\%$. Open Images is only used in ablation study.\\




\vspace{-0.5cm}
\subsection{Implementation Details}
We implement our method based on PyTorch \cite{paszke2019pytorch} with multiprocessing to support large-scale training and evaluation. The models are trained on 8 Tesla V100 GPUs. Batch size is 40 per card for training. Following \cite{ufo}, we use VGG-19 \cite{simonyan2014very} pre-trained on ImageNet \cite{deng2009imagenet} as backbone, as it achieves the best performance among different networks. All foreground objects are padded with white pixels as square images. The original object in background image is covered with rectangle mask with average value. Both foreground and background images are resized to $224\times 224$ and normalized with average value before feed into networks. We use Adam \cite{kingma2014adam} optimizer with learning rate of 0.00008 based on linear scaling rule \cite{goyal2017accurate} of multi-card training. The margins of Eq. \ref{eq:1}, \ref{eq:2} are 
set as $0.3$, $0.1$. 

\subsection{Evaluation Metrics}
For retrieval, we use the most widely used metric mAP (mean Average Precision) and R@k (Recall@k). We also report mAP-100 which is the mAP for top 100 retrievals as a fair comparison with constrained methods, because constrained retrieval methods do not rank object in all categories. When there is only one ground-truth (the original object) for each background query, we report R@k as the percentage of background queries whose ground-truth foreground reference appears in top k retrievals. Although other objects may also be compatible, the original one should always have a good rank among all the objects.\\ 

\vspace{-0.5cm}
\subsection{Comparison with Previous Methods}
In this section, we compare our method with previous works on CAIS \cite{cais} and Pixabay dataset. \cite{li2020interpretable,wu2021fine} are infeasible for comparison because they both have very different settings, and their codes and datasets are not released. \textbf{More qualitative results are included in supplementary materials}.\\
\begin{table}[!htbp]
\vspace{-1.0cm}
\caption{Comparison with previous works on CAIS in terms of mAP-100. ``\dag" denotes constrained methods with multiple models, which are not scalable in practice.}
\small
\centering
\begin{tabular}{l|c c c c c c c c|c}
\hline

\hline
Method & Boat & Bottle & Car& Chair & Dog & Painting & Person & Plant & Overall\\
\hline
Shape \cite{cais} & 7.47 & 1.16 & 10.40 & 12.25 & 12.22 & 3.89 & 6.37 & 8.82 & 7.82\\
RealismCNN \cite{zhu2015learning}& 12.33 & 7.19 & 7.55 & 1.81 & 7.58 & 6.45 & 1.47 & 12.74 & 7.14\\
\dag CFO-C Search \cite{cais}& 57.48 & 14.24 & 18.85 & 21.61 & 38.01 & 27.72 & \textbf{47.33} & 20.20 & 30.68\\
\dag CFO-D Search \cite{cais}& 55.48 & 8.93 & 24.10 & 18.16 & \textbf{57.82} & 21.59 & 27.66 & 23.13 & 29.61\\
UFO Search \cite{ufo} & 59.73 & \textbf{21.12} & 36.63 & 19.27 & 36.51 & 25.84 & 27.11 & 31.19 & 32.17\\
\hline
Ours& \textbf{70.58} & 19.41 & \textbf{40.22} & \textbf{24.17} & 37.81 & \textbf{28.20} & 44.72 & \textbf{34.91} & \textbf{37.50} \\
\hline

\hline
\end{tabular}
\vspace{-0.4cm}
\label{tab:main}
\end{table}\\
\noindent\textbf{CAIS}: We compare our method with previous state-of-the-art methods and their variants in Table \ref{tab:main}. ``Shape" \cite{cais} ranks all the foreground images by comparing their aspect ratio with the aspect ratio of query bounding box. ``RealismCNN" \cite{zhu2015learning} trains a discriminator to distinguish real/fake composite images by copy-paste original and irrelevant objects into a background image. The score of the discriminator measures the realism of each composite, which is used to rank all the objects. ``CFO-C Search" \cite{cais} and ``CFO-D Search" \cite{cais} are two constrained search methods evaluated in unconstrained scenarios. ``CFO-C Search" first trains a classifier to specify the category, then apply constrained retrieval only from this class. The results will be completely wrong if the classification fails. ``CFO-D Search" first apply constrained search to retrieve 100 samples from each category. Then it adopts the discriminator of ``RealismCNN" to re-rank all these retrievals by compositing with each background. Note that this is computationally expensive and not scalable if there are hundreds of classes. ``UFO Search" \cite{ufo} applies the discriminator to generate extra positive samples.\\
\indent In Table \ref{tab:main}, we show the mAP-100 per class and the overall performance. The proposed method significantly ($+5.33\%$) outperforms previous state-of-the-art methods on overall performance. It also achieves better performance than UFO \cite{ufo} on most of the categories. ``CFO-C Search" and ``CFO-D Search" may performs well on two categories, but they are both not scalable, because it is infeasible to train one model for each category or scan all objects with discriminator for each background on large-scale datasets with hundreds of categories. The results indicate that geometry and lighting awareness can help the retrieval in general unconstrained retrieval.\\
\begin{table*}[!htbp]
\vspace{-1.0cm}
\small
\caption{Retrieval performance of selected classes in terms of Recall@10 (\%) on Pixabay.}
\centering
\resizebox{\linewidth}{!}{
\begin{tabular}{l|cccc|cccc|c}
\hline

\hline
\multirow{4}{20pt}{Category}& \multicolumn{4}{c}{\textbf{Majority}} & \multicolumn{4}{c}{\textbf{Medium}} & \textbf{Minority}\\
& \multicolumn{4}{c}{$\#$\small{samples}$ \approx5000$}&\multicolumn{4}{c}{$\#$\small{samples}$\approx50$} & $\#$\small{samples}$<5$\\
\cline{2-10}
~ & \multirow{2}{*}{\small{Person}} & \multirow{2}{*}{\small{Flower}} & \multirow{2}{*}{\small{Birds}} & \multirow{2}{*}{\small{Vehicle}} & \small{Cell}  & \small{Mandarin}  & \small{Christmas}  & \small{Boiled} & \small{Last 50} \\
& & & & & \small{Phone} & \small{Orange} & \small{Tree} & \small{Egg}  & \small{Classes}\\
\hline
UFO \cite{ufo} & 3.84 & 8.00 & 6.74 & 8.41 & 5.71 & 36.36 & 2.94 & 6.06 & 8.00 \\
Ours &  \textbf{19.36} & \textbf{28.55} & \textbf{21.11} & \textbf{26.83}  & \textbf{20.00} & \textbf{63.64} & \textbf{20.59} & \textbf{30.30} & \textbf{24.00}\\
\hline

\hline
    \end{tabular}}
\vspace{-0.2cm}
\label{tab:pixabay}
\vspace{-0.2cm}
\end{table*}



\begin{table}[!htbp]
\vspace{-1.2cm}
\caption{Overall retrieval accuracy in terms of Recall@k (\%) on Pixabay.}
\small
\centering
\begin{tabular}{l c c c c } 
\hline

\hline
Method & R@1 & R@5 & R@10 & R@1\% \\
\hline
 UFO \cite{ufo} & 2.04 & 6.66 & 10.24 &  61.76 \\
Ours &  \textbf{7.75} & \textbf{20.13} & \textbf{28.20} & \textbf{85.61} \\
\hline

\hline
\end{tabular}
\vspace{-0.1cm}
\label{tab:large}
\vspace{-0.4cm}
\end{table}
\noindent\textbf{Pixabay}: Since none of the previous works conduct experiments on large-scale open-world setting, we implement the state-of-the-art unconstrained search method (UFO \cite{ufo}) on Pixabay and provide detailed comparison. Table \ref{tab:pixabay} shows the Recall@10 on selected categories in Pixabay. We select 4 majority and 4 medium classes, with about $5000$ and $50$ samples per class in evaluation set (Note that training set is 9 times larger). Then we compute the average Recall@10 on the last $50$ long-tail classes. The proposed method significantly outperforms UFO \cite{ufo} on all the classes. 
We also show the overall R@k on Pixabay in Table \ref{tab:large}. The proposed method significantly outperforms UFO, which means the proposed method generalizes well on large-scale open-world setting. \\
\vspace{-0.5cm}
\subsection{Ablation Study}
\label{sec:ablation}
\noindent\textbf{Geometry and Lighting Sensitivity:} Since our goal is to gain awareness on geometry and lighting, a good model should be discriminative/sensitive to the change of geometry and lighting on foreground images. We conduct experiment on Pixabay \cite{pixabay} and Open Images \cite{openimages} to verify this point. We randomly select $2,000$ foregrounds along with their corresponding background images. Then for each foreground, we use the geometry and lighting transformations in Sec. \ref{sec:contrastive} to generate transformed image. In total, we generate 100 transformed images for each foreground image, $50/50$ for geometry/lighting transformation. Then we rank the original one along with the 50 geometry or lighting transformed images to compute the Recall@k in Table \ref{tab:light_geometry}. We also measure the discriminative ability as the sensitivity to these transformations, i.e. the square euclidean distance between normalized embedding features of the original and transformed foregrounds. With L2 normalization, the square euclidean distance is $d = 2-2s$, where $s$ is the cosine similarity. Therefore higher sensitivity value means larger distance between the features of original and transformed objects, thus indicates stronger ability on distinguishing geometry or lighting transformations.\\
\begin{table*}[!htbp]
\vspace{-1.0cm}
\caption{Evaluation of sensitivity (discriminative ability) to lighting and geometry transformation. The ``sensitivity" denotes the squared L2 distance between the normalized embedding of original and transformed foreground images.}
\small
\centering
\begin{tabular}{l c c c c c c c c c}
\hline

\hline
\multirow{2}{12pt}{Ablation}& \multicolumn{4}{c}{\textbf{Lighting}} & & \multicolumn{4}{c}{\textbf{Geometry}}\\
\cline{2-5} 
\cline{7-10}
 & \small{Sensitivity($\uparrow$)} & R@5 & R@10 & R@15 & & \small{Sensitivity($\uparrow$)} & R@5 & R@10 & R@15\\
\hline
\hline
&\multicolumn{9}{c}{\textbf{Pixabay}}\\
\hline
UFO \cite{ufo} & 0.27 & 53.70 & 67.90 & 75.50 & & 0.39 & 58.25 & 69.90 & 77.95 \\
Baseline & 0.27 & 53.40 & 67.30 & 75.35 & & 0.39 & 58.10 & 69.70 & 77.65\\ 
No Contrastive &  0.51 & 55.60 & 70.70 & 79.70 & &  0.72 & 61.30 & 74.30 & 82.75\\
Overall &  \textbf{0.57} &  \textbf{60.55} & \textbf{74.70} & \textbf{82.85} & & \textbf{1.12} & \textbf{98.55} & \textbf{99.45} & \textbf{99.70}\\
\hline
&\multicolumn{9}{c}{\textbf{Open Images}}\\
\hline
Baseline &    0.24 & 51.70 & 65.45 & 74.60 & & 0.40 &  60.10 & 71.80 & 78.45\\ 
No Contrastive &   0.53 & 54.80 & 71.55 & 80.10 & &  0.98 & 71.70 & 82.60 & 89.10\\
Overall & \textbf{0.56} & \textbf{59.35} & \textbf{73.90} & \textbf{81.80} & & \textbf{1.58} & \textbf{99.50} & \textbf{99.75} & \textbf{99.90}\\
\hline

\hline
\end{tabular}
\vspace{-0.1cm}
\label{tab:light_geometry}
\end{table*}
\begin{figure*}[!htbp]
    \vspace{-1.5cm}
    \centering
    \includegraphics[width=0.99\linewidth]{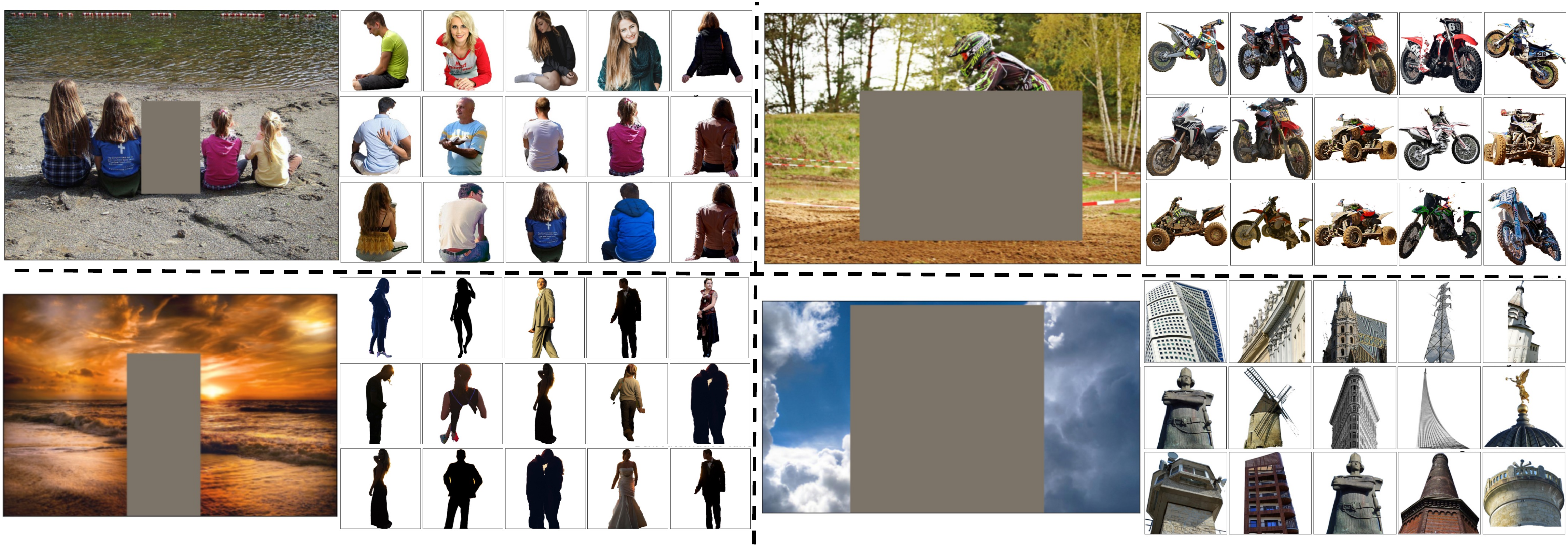}
    \vspace{-0.3cm}
    \caption{Qualitative comparison on Pixabay and Open Images. Each of the four examples contains three rows of retrieval results, corresponding to: ``Baseline" (1st row), ``No Contrastive" (2nd row), ``Overall" model (3rd row). }
    \label{fig:qualitative}
\end{figure*}\\
\indent In Table \ref{tab:light_geometry}, the overall proposed method has much higher sensitivity to both geometry and lighting transformations for both datasets. UFO \cite{ufo} and ``Baseline" both use fixed foreground features, thus have a low sensitivity. ``No Contrastive" removes the proposed contrastive transformation but keeps the alternating training, which has a lower sensitivity and retrieval performance than ``Overall", but better than ``Baseline". The results demonstrate the effectiveness of both alternating training and contrastive learning on improving discriminative ability. We also show qualitative results of our ablations in Fig. \ref{fig:qualitative} with diverse scenes, viewpoints and lighting. Our overall method better respects the geometry and lighting than baseline methods.\\
\noindent\textbf{Components Ablations:} We further conduct detailed ablation study on CAIS to show the effectiveness of each component.
``Direct Training" mean directly training two networks with loss in Eq. \ref{eq:1}. ``Aug" denotes direct training along with mask augmentations in Sec. \ref{sec:alternative} to prevent the model from overfitting on edge pixels. ``Fix+Aug" is also used as ``Baseline" in other tables. It uses a fixed 
\begin{wraptable}{r}{0.4\linewidth}
\vspace{-0.5cm}
\caption{Ablation study on components.}
\small
\centering
\begin{tabular}{l|c c}
\hline

\hline
Ablations & mAP & mAP-100 \\
\hline
Direct Training & 17.93  & 24.02 \\
Aug & 23.45 & 31.78 \\
Fix+Aug & 28.65 & 30.33 \\
No Alternating & 29.99 & 32.13 \\
No Contrastive & 31.20 & 36.30 \\
\hline
Overall & \textbf{32.67} & \textbf{37.49} \\
\hline

\hline
\end{tabular}
\vspace{-0.3cm}
\label{tab:ablation}
\end{wraptable}
foreground network with mask augmentations. Our ``Overall" method adopt both alternating training and contrastive learning with mask augmentations. ``No Alternating" removes alternating training from ``Overall", and ``No Contrastive" removes the contrastive learning loss in Eq. \ref{eq:2}. As shown in Table \ref{tab:ablation}, ``Overall" performs the best among all the ablations, indicating the effectiveness of each component.\\

\hspace{-0.5cm}
\begin{minipage}{0.45\linewidth}
\vspace{-0.2cm}
\captionsetup{justification=centering}
\captionof{table}{Ablation study on different \\ alternating training strategies.}
\small
\centering
\begin{tabular}{l|c c}
\hline

\hline
Ablations & mAP & mAP-100 \\
\hline
Zero Round & 28.65 & 30.33 \\
One Round & \textbf{31.20} & \textbf{36.30} \\
Two Rounds & 30.35 & 36.22 \\
Reverse Order & 28.82 & 36.14 \\
\hline

\hline
\end{tabular}
\label{tab:ablation_alternative}
\end{minipage}
\hspace{0.2cm}
\begin{minipage}{0.45\linewidth}
\captionof{table}{Ablation study on different transformations.}
\small
\centering
\begin{tabular}{l|c c}
\hline

\hline
Ablations & mAP & mAP-100 \\
\hline
No Contrastive & 31.20 & 36.30 \\
Geometry & 32.11 & 36.68\\
Geometry+Color & 30.80 & 35.06 \\
Geometry+Lighting & \textbf{32.67} & \textbf{37.49} \\
\hline

\hline
\end{tabular}
\label{tab:ablation_transformation}
\end{minipage}
\vspace{0.2cm}

\noindent\textbf{Training Strategy Ablations:} We conduct ablation study on different alternating training strategies and select the best strategy for our overall method. All these ablations adopt mask augmentations without using contrastive learning. ``Zero Round" means no alternating training, which is equivalent to the ``Baseline". ``One Round" ($\operatorname*{argmin}_{N_{b}} \mathcal{L} $, then $\operatorname*{argmin}_{N_{f}} \mathcal{L} $) trains foreground network after the background network is trained. ``Two Round" trains the background network for another round based on the ``One Round" model. ``Reverse Order" ($\operatorname*{argmin}_{N_{f}} \mathcal{L} $, then $\operatorname*{argmin}_{N_{b}} \mathcal{L} $) first trains the foreground network, then trains the background network. Table \ref{tab:ablation_alternative} shows that ``One Round" strategy performs the best among all the strategies, indicating that training more rounds is not beneficial. We thus use one round for our overall method.\\
\noindent\textbf{Transformation Ablations:} To show the effectiveness of each transformation, we train our method with different transformations in contrastive learning. ``No Contrastive" removes contrastive learning loss in Eq. \ref{eq:2}. ``Geometry" only adopts geometry transformation in Fig. \ref{fig:transformation}, and ``Geometry+Lighting" means each foreground is applied with both transformations. ``Geometry+Color" uses linear color jittering instead of our non-linear lighting transformation on the top of ``Geometry". In Table \ref{tab:ablation_transformation}, the ``Geometry+Lighting" outperforms other ablations, indicating that both transformations are necessary in our framework, and linear color jittering does not well simulate lighting changes. 




\vspace{-0.3cm}
\subsection{Location and Scale Prediction}
\begin{figure*}[!htbp]
    \centering
    \vspace{-1.1cm}
    \includegraphics[width=.75\linewidth]{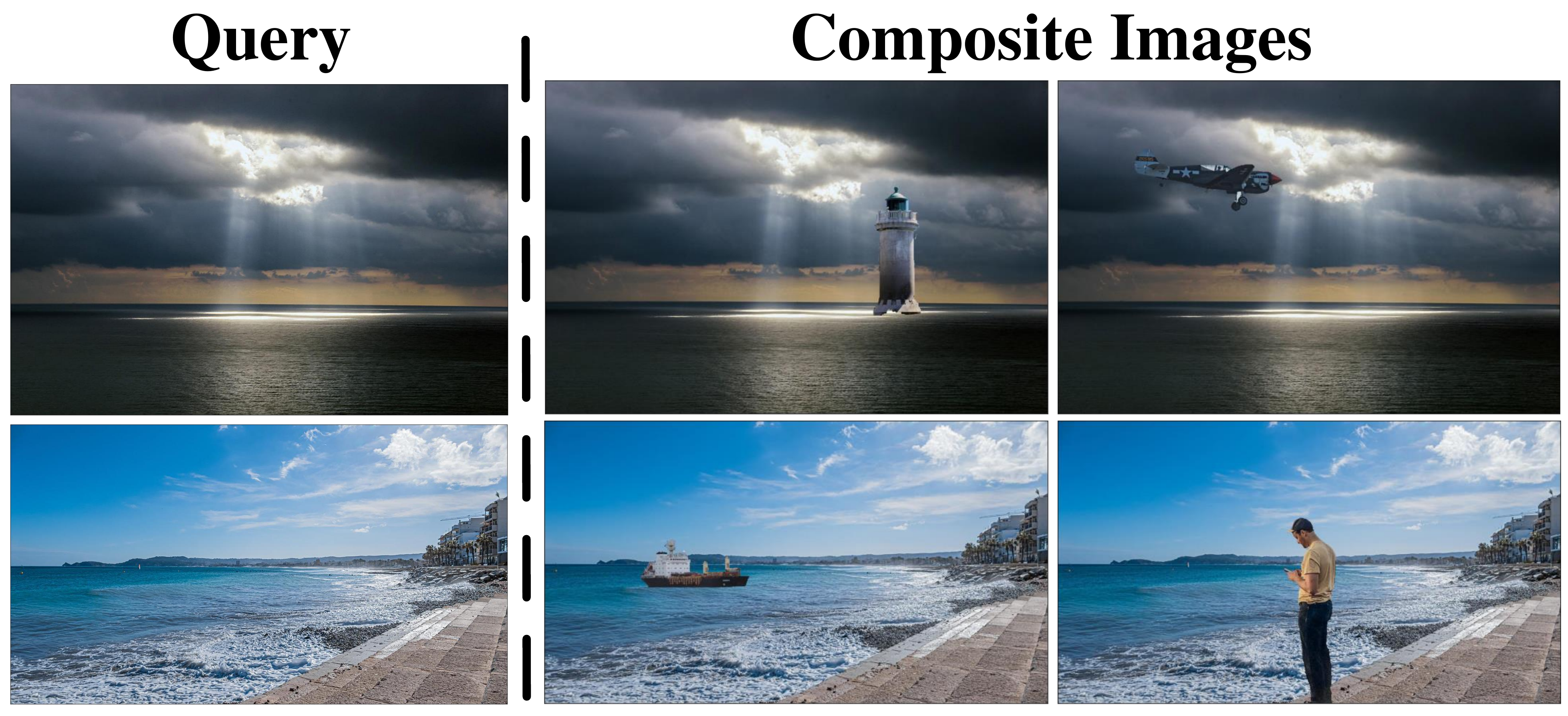}
    \vspace{-0.2cm}
    \caption{Qualitative results of composite images for non-box scenarios, generated based on top retrieved objects with automatically predicted location/scale and harmonization \cite{Jiang_2021_ICCV}. }
    \label{fig:qualitative-final}
    \vspace{-0.5cm}
\end{figure*}

\begin{figure}
    \centering
    \vspace{-0.7cm}
    \includegraphics[width=.8\textwidth]{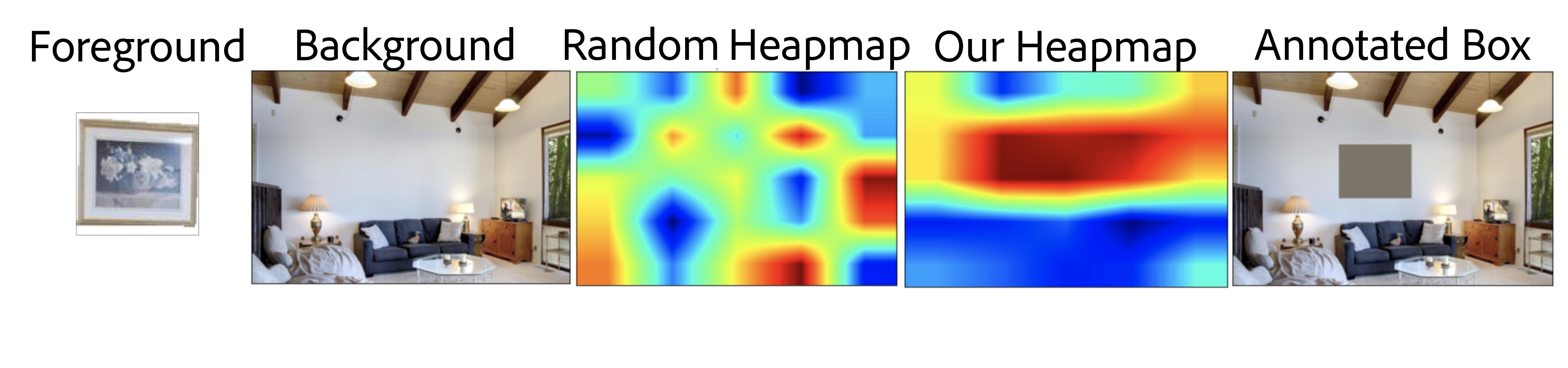}
    \vspace{-0.7cm}
    \caption{Example heatmaps for location prediction evaluation.}
    \label{fig:heatmap}
    \vspace{-0.5cm}
\end{figure}

Since CAIS has annotated bounding boxes on intact background images, we evaluate our location and scale prediction method based on these boxes for non-box scenarios. We compare the proposed method with random strategy and ``Baseline" model. For each background image, we randomly select $5$ of its annotated compatible foregrounds for evaluation and compute the average value to report. Qualitative results are included in Fig. \ref{fig:qualitative-final}.\\
\\
\indent For location prediction, given a background along with its annotated foreground object, we first apply $10\time 10$ grid search with sliding window method in Sec. \ref{sec:location}. To evaluate location and scale separately, the scale here is fixed as the scale of annotated box. Then bilinear interpolation is adopted to generate the heapmap, as shown in Fig. \ref{fig:heatmap}. The heapmap value is then normalized with the maximum and minimal value in the grid matrix, so that all values are in $[0,1]$. Higher value means the model assigns high compatibility on the corresponding location. Note that the annotated box location may not be the only good location, other boxes with zero IOU (Intersection Over Union) can also have compatible locations, e.g. all the non-occluded locations on the wall in Fig. \ref{fig:heatmap}. However, the ground-truth location should always have a good compatibility score as compared with scores of other locations. Therefore, we compute the similarity score on the annotated location and normalize it with the same normalization factors as the heatmap. A good location prediction model should give a high normalized similarity (NS) on the annotated locations. For ``Random" strategy, the normalized similarity is randomly selected in $[0,1]$ since the heatmap is random. \\
\indent For scale prediction, given a background along with its annotated foreground object, all methods use the annotated ground-truth locations as the center of the box. With a fixed initial size with $1/25$ area of the whole background image, we assign $9$ different scales by $scale = 1.2^{k-4}, k=0,1,...,8$. The box size is then multiplied by the scale to get $9$ candidate boxes. ``Random" strategy selects one scale randomly, while the other two methods use the similarities between the background with boxes and the foreground to rank all the scales. We then compute the IOU between the predicted box and the annotated box. \\
\begin{table*}[!htbp]
\vspace{-1.0cm}
\caption{Evaluation on location and scale prediction. NS (normalized similarity) denotes the cosine similarity of the ground truth location normalized by the maximum and minimum similarity of the other sliding window locations.}
\small
\centering
\begin{tabular}{l c c c c c c c}
\hline

\hline
\multirow{2}{*}{Method}& \multicolumn{3}{c}{\textbf{Location}} & & \multicolumn{3}{c}{\textbf{Scale}}\\
\cline{2-4} 
\cline{6-8}
& $NS>0.99$ & $NS>0.95$ & $NS>0.9$ & &$IOU>0.9$ & $IOU>0.75$ & $IOU>0.5$\\
\hline
\hline
Random &  5.50 & 9.25 & 12.25 & & 4.25 & 14.50 & 38.25\\ 
Baseline &  15.25 & 25.25 & 38.25 & & 5.75 & 19.75 & 48.50\\ 
Ours &  \textbf{22.25} & \textbf{31.50}  & \textbf{39.00} & & \textbf{6.75} & \textbf{24.25} & \textbf{51.75}\\
\hline

\hline
\end{tabular}
\vspace{-0.7cm}
\label{tab:location}
\end{table*}
\\
\indent As shown in Table \ref{tab:location}, we report the percentage of correct predictions with different thresholds on two evaluation metrics, i.e. NS and IOU. Our method performs the best on both location and scale prediction, indicating the superiority of the proposed method on non-box scenarios. 

\section{Limitations and Potential Social Impact}
Current framework only uses 2D transform for contrastive learning, which is efficient and scalable, but not as accurate as 3D transform. This limitation could be addressed in the future by adopting 3D transform as better pre-processing w/o changing the proposed learning framework. Another limitation is that the search space is currently bounded by the gallery, and hence there may not be any perfectly compatible object images in the database even with the large-scale open-world database setting we adopt in our work. One solution is to augment the search space by allowing transformation of objects in the database.\\
\indent Retrieval-based compositing could be indirectly used to generate fake images, but we can make sure all the gallery object images do not have ethical or privacy issues. Also, fake images usually have incompatible geometry or lighting conditions which can be detected using our model.

\section{Conclusion}
We propose a novel unconstrained foreground search method for compositing (GALA), as the first to learn geometry-and-lighting awareness from large-scale data for real-world scenarios. GALA achieves state-of-the-art results on CAIS and Pixabay dataset.   
It also tackles non-box scenarios 
by automatic location-scale prediction, which is not explored by previous works.  

\clearpage
%
%
\bibliographystyle{splncs04}
\bibliography{egbib}
\end{document}